\documentclass[twoside]{article}
\usepackage{amsfonts,amssymb,amsbsy,textcomp,marvosym,amsmath,caption,threeparttable,amsthm,subfigure,float,lastpage,lscape}
\usepackage{eurosym,mathrsfs,fancyhdr,CJK,multicol,graphics,indentfirst,color,bm,upgreek,booktabs,graphicx,multirow,warpcol}
\usepackage{epstopdf}
\usepackage{wrapfig}

\usepackage{hyperref}
\newcommand\nnfootnote[1]{%
  \begin{NoHyper}
  \renewcommand\thefootnote{}\footnote{#1}%
  \addtocounter{footnote}{-1}%
  \end{NoHyper}
}

\usepackage{xspace,xpunctuate}

\newcommand{\ie}{\textit{i.e.},\xspace}
\newcommand{\etal}{\xspace\textit{et al.}\xspace}
\newcommand{\eg}{\textit{e.g.},\xspace}

\def\footnoterule{\kern 1mm \hrule width 10cm \kern 2mm}

\allowdisplaybreaks
\catcode`@=11
\def\title#1{\vspace{3mm}\begin{flushleft}\vglue-.1cm\Large\bf\boldmath\protect\baselineskip=18pt plus.2pt minus.1pt #1
\end{flushleft}\vspace{1mm} }
\def\author#1{\begin{flushleft}\normalsize #1\end{flushleft}\vspace*{-4pt} \vspace{3mm}}
\def\address#1#2{\begin{flushleft}\vglue-.35cm${}^{#1}$\small\it #2\vglue-.35cm\end{flushleft}\vspace{-2mm}\par}

\def\jz#1#2{{$^{\footnotesize\textcircled{\tiny #1}}$\let\thefootnote\relax\footnotetext{\!\!$^{\footnotesize\textcircled{\tiny #1}}$#2}}}
\catcode`@=11
\def\section{\@startsection{section}{1}{\z@}%
 {-3ex \@plus -.3ex \@minus -.2ex}%
 {2.2ex \@plus.2ex}%
{\normalfont\normalsize\protect\baselineskip=14.5pt plus.2pt minus.2pt\bfseries}}
\def\subsection{\@startsection{subsection}{2}{\z@}%
 {-3ex\@plus -.2ex \@minus -.2ex}%
 {2ex \@plus.2ex}%
{\normalfont\normalsize\protect\baselineskip=12.5pt plus.2pt minus.2pt\bfseries}}
\def\subsubsection{\@startsection{subsubsection}{3}{\z@}%
 {-2.2ex\@plus -.21ex \@minus -.2ex}%
 {1.4ex \@plus.2ex}
{\normalfont\normalsize\protect\baselineskip=12pt plus.2pt minus.2pt\sl}}

\begin{document}
\begin{CJK*}{GBK}{song}
\thispagestyle{empty}
\vspace*{-13mm}
\vspace*{2mm}

\nnfootnote{}

\title{Computational Approaches for Traditional Chinese Painting: From the ``Six Principles of Painting'' Perspective}

\author{Wei Zhang$^{1}$, Jian-Wei Zhang$^{1}$, Kam Kwai Wong$^{2}$, Yifang Wang$^{3}$, Yingchaojie Feng$^{1}$, Luwei Wang$^{1}$, and Wei Chen$^{1,4,*}$}

\address{1}{State Key Lab of CAD\&CG, Zhejiang University, Hangzhou 310058, China}
\address{2}{Hong Kong University of Science and Technology, Hong Kong 999077, China}
\address{3}{Kellogg School of Management, Northwestern University, Evanston 60208, U.S.A}
\address{4}{Laboratory of Art and Archaeology Image, Zhejiang University, Hangzhou 310058, China}

\vspace{2mm}

\noindent E-mail: zw\underline{~~}yixian@zju.edu.cn; zjw.cs@zju.edu.cn; kkwongar@cse.ust.hk; yifang.wang@kellogg.northwestern.edu; fycj@zju.edu.cn; ppwlwpp@zju.edu.cn; chenvis@zju.edu.cn \\[-1mm]

\noindent {\small\bf Abstract} \quad  
{\small {Traditional Chinese Painting (TCP) is an invaluable cultural heritage resource and a unique visual art style. In recent years, increasing interest has been placed on digitalizing TCPs to preserve and revive the culture. The resulting digital copies have enabled the advancement of computational methods for structured and systematic understanding of TCPs. To explore this topic, we conducted an in-depth analysis of 92 pieces of literature. We examined the current use of computer technologies on TCPs from three perspectives, based on numerous conversations with specialists. First, in light of the ``Six Principles of Painting" theory, we categorized the articles according to their research focus on artistic elements. Second, we created a four-stage framework to illustrate the purposes of TCP applications. Third, we summarized the popular computational techniques applied to TCPs. The framework also provides insights into potential applications and future prospects, with professional opinion. The list of surveyed publications and related information is available online at https://ca4tcp.com.}}


\vspace*{3mm}


\vspace*{4mm}

\end{CJK*}
\baselineskip=18pt plus.2pt minus.2pt
\parskip=0pt plus.2pt minus0.2pt
\begin{multicols}{2}

\section{Introduction}
\label{sec:Introduction}
\begin{figure*}[!htb]
\centering
\includegraphics[width=17.1cm,height=5cm]{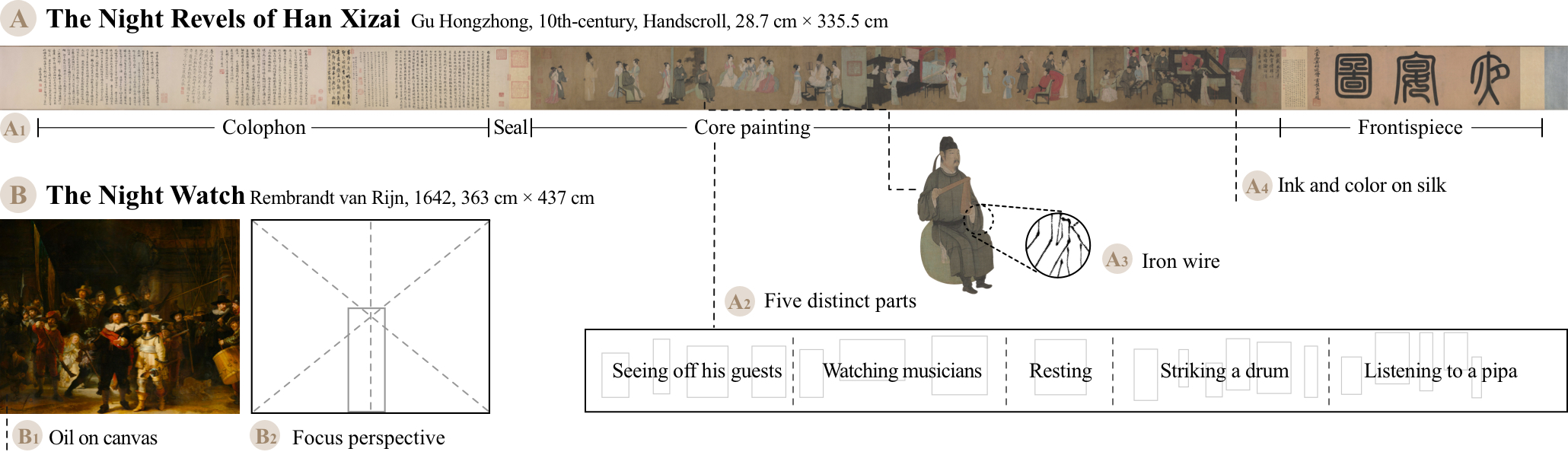}
\caption{``The Night Revels of Han Xizai"~\cite{hxz2022painting} mainly depicts the scene of a night banquet hosted by Han Xizai (A), the prime minister of the Tang Dynasty, while ``The Night Watch"~\cite{nw2022painting} depicts the scene of a night patrol by the civic militia (B). Both paintings depict group portraits in terms of their subject matter, depicting dozens of characters. However, there are significant differences in terms of the materials, format, and techniques used in the paintings.}
\label{fig:tcp_w}
\end{figure*}

Originating from the Han Dynasty, Traditional Chinese Painting (\textbf{TCP}) has been a primary art form in China~\cite{cheng2018essential}, characterized by artistic expressions depicted on paper and silk with brushes dipped in black ink and Chinese pigments.
TCP has been used to express the author's artistic creativity and to insinuate criticism of the society, philosophy, and politics of the time (\autoref{fig:tcp_w}A). 
Existing studies on TCP mainly focus on its history, explanation, and appreciation of paintings, as well as various styles and techniques. 
However, these are typically conducted thorough a close-reading and case study approach based on painting theories~\cite{bradley2018visualization}, which while insightful, is time-consuming and unscalable for studying the patterns and evolution trends of TCP that have emerged over the centuries.

\baselineskip=18pt plus.2pt minus.2pt
\parskip=0pt plus.2pt minus0.2pt

Recent advances in computer technology have greatly improved the efficiency of TCP research. For example, Convolutional Neural Networks (CNNs) detect and segment the elements in TCP~\cite{8419282,zhang2011multispectral,chen2012simulating}, while Generative Adversarial Networks (GANs) generate TCPs and perform style transfer~\cite{xue2021end,zhang2020detail,he2018chipgan,li2021immersive,9413063}. However, applying these deep learning techniques on TCP requires a comprehensive understanding of their characteristics. For instance, although AI-generated TCPs may resemble the originals when viewed from afar, the brushstroke details can be far from natural.
Moreover, the visual objects are often mislocated with regard to the TCP composition style. Oil paintings have similar challenges and have been studied with regard to their specific characteristics, such as stroke composition~\cite{zheng2018strokenet,litwinowicz1997processing,huang2019learning,zou2021stylized,kotovenko2021rethinking,liu2021paint}.
Nevertheless, such analyses and conclusions cannot be directly applied to TCPs, since they differ greatly in terms of material, format, and techniques (see \autoref{tbl:tcp_oil}). 
They have different requirements for ink diffusion effects, point of view considerations, and painting techniques to depict textures.

\begin{table*}[tb]
    \renewcommand\arraystretch{1.5}
	\centering 
	\caption{Differences between traditional Chinese painting and oil painting.}  %
	\label{tbl:tcp_oil}
	\begin{tabular}{p{1.5cm}|p{7cm}|p{7.6cm}}
		\toprule
		\textbf{Aspects} & \textbf{Traditional Chinese painting} & \textbf{Oil painting} \\
		\midrule
        Material   & Painting black ink and Chinese pigments (similar to gouache paint) on paper and silk (\autoref{fig:tcp_w}A4). & Applying the mixture of pigments and drying oils with diverse plasticity on wood and canvas (\autoref{fig:tcp_w}B1).\\
        Format    & Contain several sections (\autoref{fig:tcp_w}A1); The point of views can come from different scenes (\autoref{fig:tcp_w}A2).  & Depict the scene from a focal perspective (\autoref{fig:tcp_w}B2).\\
        Technique & Objects are depicted by different brushwork systems, \eg wrinkling techniques for landscapes and calligraphic-line techniques for figures. (\autoref{fig:tcp_w}A3). & Focus on realism, using light and dark tones to express the texture of objects (\autoref{fig:tcp_w}B).\\
		\bottomrule
	\end{tabular}
\vspace{-0.2in}
\end{table*}

Various papers on computer vision techniques have been conducted mainly from the perspectives of aesthetic judgment and stylization.
DiVerdi~\cite{diverdi2015modular} investigated the modular framework for digital paintings and loosely addressed TCP by calligraphy.
Zhang\etal\cite{zhang2021comprehensive} surveyed systems of photographs and paintings from the perspective of aesthetic evaluation.
Kyprianidis\etal\cite{kyprianidis2012state} reviewed methods for transforming photos into aesthetically stylized renderings.
Li\etal\cite{li2022computing} conducted a review of the computer methods used during various stages of production and preservation of Chinese cultural heritage.
TCP, however, is only viewed as a common format of image data in these papers, which overlook the peculiarities of Chinese paintings' data attributes and analysis tasks.

To bridge the gap between general image data and TCP, we conducted a systematic review of the literature on key areas such as data visualization (VIS), Computer Vision (CV), Computer Graphics (CG), and Human-Computer Interaction (HCI).
We drew inspirations from the TCP appreciation theory and adapted the ``Six Principles of Painting" to modern concepts for categorizing the collected literature (\autoref{sec:background}). 

In collaboration with Chinese painting specialists, we proposed an four-stage framework to review the purposes of using computational techniques in TCP (\autoref{sec:ana_fmwk}). 
We then classified the recent computer-aided techniques from the perspectives of task, feature, and rendering (\autoref{sec:tech}). 
Lastly, we reported the discussions with specialists about the current drawbacks and potential future applications of computer technology to TCP (\autoref{sec:challenge}). We believe this paper can offer an explanation, insights, and examples into every aspect of TCP, thus making way for more comprehensive appreciation. An interactive browser of this paper is available at https://ca4tcp.com.

\section{Methodology}
\label{sec:meth}

This section describes the search methodology, corpus construction, and collaboration with experts.

\subsection{Methods and Corpus}

We constructed a literature corpus based on keyword- and relation-search methods. We focused on TCP-related keywords (\eg ``Chinese landscape painting'', ``Chinese ink wash painting'', and ``Chinese brush painting''), resulting in an initial 88 papers.
To enlarge our literature corpus, we further used the relation-search method. We identified eight influential papers from the initial corpus and exhaustively traversed their references and citations, which expanded the corpus to 112 papers.
We thoroughly assessed the corpus based on relevance, focusing on publications that probe methodologies and applications. Papers on theory~\cite{wu2013modeling} and evaluation~\cite{bo2018computational} were excluded. 
Despite extensive research on brushes, we only included those relevant to TCP, discarding the calligraphy-related articles~\cite{mi2002droplet, bai2007efficient}. Finally, the corpus comprised 92 papers.

\subsection{Cooperate with Domain Expert}

Over the past year, we have been working closely with two experts to strengthen our understanding of the specialized and unique domain of TCP.
The experts include a professor with over 20 years of expertise in Chinese painting, and a doctoral candidate in Chinese painting theory with five years of experience.
Our collaboration consists of the following stages:
Firstly, we consulted the experts about the domain knowledge of TCP and the domain interpretation of some exemplar papers.
Secondly, we iteratively refined the analysis framework for the application of computer technology to TCP.
Finally, we explored research challenges and opportunities by discussing the findings and proposing future research projects.

\subsection{Coding and Classification}

Through iterative discussions with experts, we have analyzed the corpus from three perspectives: research scope in TCP~\autoref{sec:background}, specifically-targeted problem~\autoref{sec:ana_fmwk}, and the use of computer-based methods~\autoref{sec:tech}. 
To differentiate the TCP research from image data analysis properly, the characteristics of TCP should be taken into account. 
We noticed that the ``Six Principles of Painting'' has summarized the most important considerations for drawing and appreciating TCP. 
We used it as the coding scheme for categorizing the literature with different targeted problems, \ie the concerned artistic elements. 
In addition, we evaluated the current analysis of TCP, and concluded with a paradigm which outlines the components and purposes of TCP analysis that are supported by modern computer techniques. 
Specifically, the paradigm includes digitalization, interpretation, creation, and exhibition of TCP.
Lastly, we coded the papers in the corpus according to the types of computational techniques they utilized, rather than the concrete algorithms that are largely interchangeable.

During the paper analysis, three authors independently coded 92 papers over a period of four weeks (\autoref{fig:code}). 
The classification criteria were refined during the coding process. 
In cases where there were disputes regarding categorization, all authors were involved in debates to reach a consensus.
For instance, we first restricted the classification of the papers using the ``Resonance of the Spirit'' in the Six Principles to the subjective evaluation of static Chinese painting images. 
After deliberation, we decided that this idea should be expanded to include animations, as this was deemed to be a more expressive art form which could be evaluated by the principle of ``Resonance of the Spirit''. More discussions are given in \autoref{sec:spirit}.

\setcounter{figure}{1}
\begin{figure*}[bp]
\centering
\includegraphics[height=\textheight]{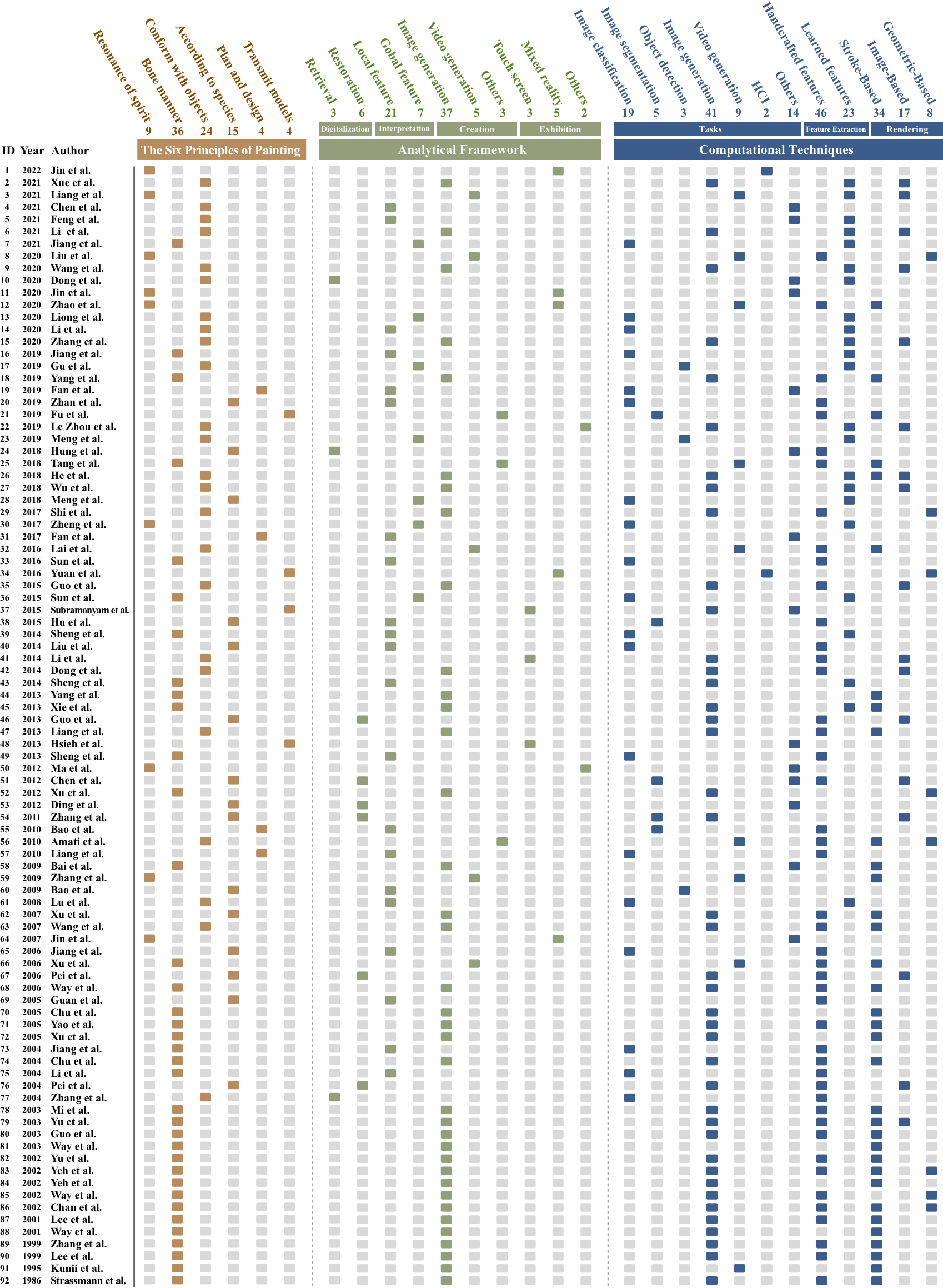}
\caption{The 92 collected papers and their codes.}
\label{fig:code}
\end{figure*}

\section{The Six Principles of Painting}
\label{sec:background}
The \textit{Six Principles of Painting} were proposed in the sixth century to serve as the grading standards of TCP~\cite{lin1967art}.
They have remained influential to this day and shaped the ways in which TCPs are drawn and appreciated.
We selected the translations collected in~\cite{van1962way} to clarify these principles in the following sections.

\subsection{Resonance of the Spirit, Movement of Life}
\label{sec:spirit}

\hspace{0.1pt}
\vspace{-25pt}
\begin{wrapfigure}[2]{l}{0.02\textwidth}
 \begin{center}
  \vspace{-25pt}
  \includegraphics[width=0.057\textwidth]{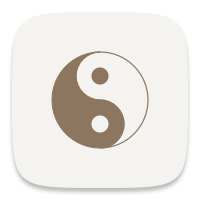}
 \end{center}
\end{wrapfigure}
\noindent
The first principle emphasizes the delivery of vitality in the objects and emotions.
This principle is considered the most important, upon which the remaining principles were developed~\cite{lin1967art}.
While the resonated spirits and lively movements are difficult to evaluate objectively, we adopted their semantic meaning and defined them as techniques that aim to induce emotional arousal and engage audiences.

Papers within this category analyze the conveyed emotions and use this information to recreate captivating paintings on various devices.
For example, Zheng\etal~\cite{zheng2017chinese} utilized machine learning methods to extract relevant features and classified TCPs with the conveyed emotions.
To enhance the emotional expressions, other works~\cite{lianginstance,liu2020animating,zhang2009video} attempted to make the objects in the paintings `alive' (\eg animations), so that the paintings can be interpreted vividly.
In recent years, significant amount of efforts~\cite{jin2022immersive,jin2020reconstructing,zhao2020shadowplay2,ma2012annotating,jin2007real} have utilized mixed reality technology to display TCPs, giving viewers new perspectives on appreciating the paintings.

\subsection{Bone Manner, Structural Use of the Brush}
\hspace{0.1pt}
\vspace{-25pt}
\begin{wrapfigure}[2]{l}{0.02\textwidth}
 \begin{center}
  \vspace{-25pt}
    \includegraphics[width=0.057\textwidth]{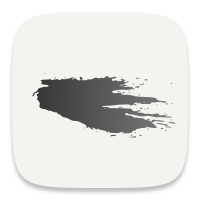}
 \end{center}
\end{wrapfigure}
\noindent
The bone method corresponds to the use of the brush.
Calligraphy and paintings highly influenced one other, with each brush stroke having its own structure, texture, and meaning.
We defined this principle as rendering techniques that involve brush strokes.
There are 36 articles discussing the bone method from the following aspects:

\emph{Stroke extraction.} Some articles focus on identifying and replicating the distinctive brushstrokes of TCP~\cite{chan2002two,jiang2021mtffnet,li2004studying,sun2016monte,sun2015brushstroke,sheng2014,sheng2014recognition,sheng2013style,xu2006animating,yeh2002non,yu2003image}. 
Frequent subjects in TCP, including mountains~\cite{way2002synthesis}, rocks~\cite{way2001synthesis}, and trees~\cite{zhang1999simple}, were imitated.

\textbf{Ink simulation.} TCPs are drawn with brushes dipped in black ink or Chinese pigments, which diffuse on rice papers or silk. Therefore, the simulation of ink effect is heavily studied~\cite{xu2012stroke,way2006computer,10.1145/1073204.1073221,chu2004real,mi2004droplet,guo2003nijimi,way2003physical,yu2002model,lee2001diffusion,kunii1995diffusion,10.1145/15886.15911}. On the other hand, the physical model of the brush itself also deserves in-depth analysis~\cite{bai2009chinese,xu2005virtual,yeh2002effects,lee1999simulating}.

\textbf{Style recognition.} Xieyi and Gongbi are two representation styles of TCP. Xieyi uses techniques that privilege the spontaneity of the line (\hyperref[fig:gongbixieyi]{Fig.~3}A). 
Gongbi uses highly detailed brushstrokes that precisely delimit details (\hyperref[fig:gongbixieyi]{Fig.~3}B). 
These stroke characteristics were used to classify different painting styles~\cite{jiang2019dct,yang2019easy,jiang2004categorizing}.

\textbf{Painting process reproduction.} Understanding the painting process of TCP is particularly useful for painting practice and education~\cite{8113507,yang2013animating,xie2013artist,yao2005painting}. It also provides a practical basis for painting generation.

\subsection{Conform with the Objects, Obtain their Likeness}
\hspace{0.1pt}
\vspace{-25pt}
\begin{wrapfigure}[2]{l}{0.02\textwidth}
    \begin{center}
    \vspace{-25pt}
    \includegraphics[width=0.057\textwidth]{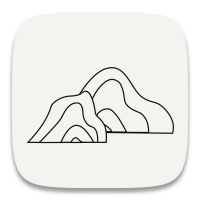}
    \end{center}
\end{wrapfigure}
\noindent
Chinese painters have developed a distinct way of depicting objects in the world.
Compared with their Western counterparts, objects in Chinese paintings are more surreal.
This principle is reinterpreted as extracting objects from TCP and depicting and sketching natural objects in the TCP artistic style.
Six articles~\cite{dong2020feature,liong2020automatic,li2020multi,gu2019deep,lu2008content,zhang2004modelling} have studied the TCP classification according to the extracted objects, and two articles~\cite{chen2021poemgeneration,feng2022ipoet} produce text descriptions (\ie instance-level captions) for the objects in TCP. There are additional 16 articles that conduct the style transfer for natural objects with TCP art styles~\cite{xue2021end,li2021immersive,9413063,zhang2020detail,le2019walking,meng2019elements,he2018chipgan,wu2018research,shi2017generative,lai2016data,guo2015novel,li2014writing,dong2014real,liang2013image,amati2010modeling,wang2007image}.

\setcounter{figure}{2}
\begin{center}
\includegraphics[width=8.2cm]{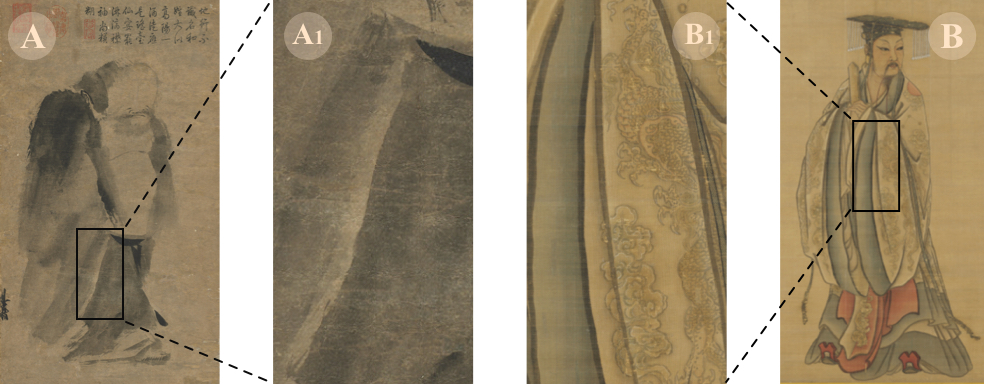}\\
\vspace{3mm}
\parbox[c]{8.3cm}{\footnotesize{Fig.3.~}  Liang Kai's ``Immortal in Splashed Ink" (A)~\cite{xy2022painting} exemplifies the Xieyi style, using wet strokes of monochromatic ink to create the immortal's cloth (A1), while Ma Lin's ``King Yu of Xia" (B)~\cite{gb2022painting} exemplifies the more elaborate Gongbi style for clothing patterns (B1).}
\label{fig:gongbixieyi}
\end{center}

\subsection{According to the Species, Apply the Colors}
\hspace{0.1pt}
\vspace{-25pt}
\begin{wrapfigure}[2]{l}{0.02\textwidth}
 \begin{center}
  \vspace{-25pt}
  \includegraphics[width=0.057\textwidth]{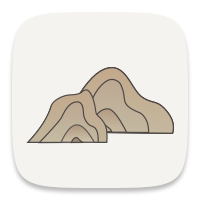}
 \end{center}
\end{wrapfigure}
\noindent
The suitability to type often appears to judge the correct use of colors.
In the drawing process, inks have to be applied in multiple layers to arrive at the desired tone.
Since inks and papers react to the environment and receive damages, ancient paintings have lost their original states and need to be conserved properly.
This principle corresponds to color analysis and restoration of paintings.
Papers that fall into this category concentrate on understanding the object classes~\cite{zhan2019,hung2018study,meng2018classification,liu2014Classification,bao2009effective,guan2005automatic}
and performing further actions (\eg color enhancement~\cite{hu2015object,xu2007generic,jiang2006effective}). 
Additionally, understanding the color of TCP is crucial for repairing them. Some papers employ computer technologies to address the issues of deteriorating paper and fading color~\cite{guo2013image,chen2012simulating,zhang2011multispectral,pei2006background,pei2004virtual,ding2012research}.

\subsection{Plan and Design, Place and Position}
\hspace{0.1pt}
\vspace{-25pt}
\begin{wrapfigure}[2]{l}{0.02\textwidth}
 \begin{center}
  \vspace{-25pt}
  \includegraphics[width=0.057\textwidth]{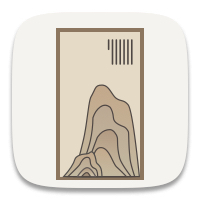}
 \end{center}
\end{wrapfigure}
\noindent
Division and planning refer to the positioning and arrangement of objects.
A unique layout characteristic in Chinese painting is the concept of ``void'' (white space).
It is believed that leaving some part of the paper blank could induce more imagination in readers' minds~\cite{fan2019evaluation,fan2017visual}.
The seals, preface, and postscript are also significant components of the composition of TCP~\cite{bao2010novel,liang2010simple}.
Papers in this category concentrate on analyzing and enhancing the composition.

\subsection{To Transmit Models by Drawing}
\hspace{0.1pt}
\vspace{-25pt}
\begin{wrapfigure}[2]{l}{0.02\textwidth}
 \begin{center}
  \vspace{-25pt}
  \includegraphics[width=0.057\textwidth]{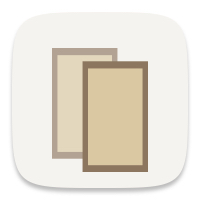}
 \end{center}
\end{wrapfigure}
\noindent
Prior to the invention of printers, the manual replication of paintings was required to facilitate their distribution and use as commodities.
Copying the classics and antique masterpieces also help amateurs improve their skills by closely observing the techniques.
Papers in this area has concentrated on digitalizing Chinese paintings and showing them on different devices, such as high relief~\cite{8419282}, virtual reality reconstruction~\cite{yuan2016tunable}, and interactive devices~\cite{subramonyam2015sigchi,hsieh2013viewing}.
The primary distinction between this principle and others is that these methods focus on precise reproduction and minimal modification from the original copies.
\section{Analytical Framework}
\label{sec:ana_fmwk}

In this section, we propose a framework for applying computational techniques in TCP based on the state-of-the-art and the expertise of domain experts. 
As shown in \autoref{fig:datapipeline}, the framework involves four typical stages, from the \textbf{digitalization} and \textbf{interpretation} of existing paintings to the \textbf{creation} of new artworks. 
These three stages will also serve the purpose of \textbf{exhibition}.

\setcounter{figure}{3}
\begin{figure*}[!htb]
\centering
\includegraphics[width=17.4cm,height=2.5cm]{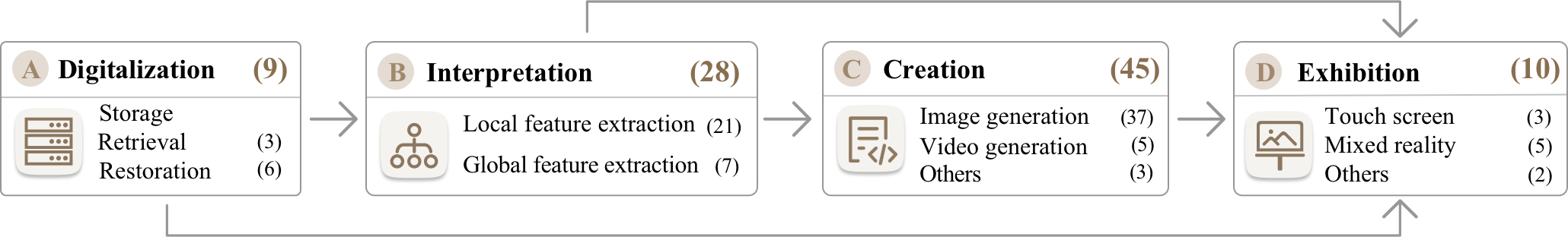}
\caption{A framework for applying computational techniques in TCP analysis.}
\label{fig:datapipeline}
\end{figure*}
\baselineskip=18pt plus.2pt minus.2pt
\parskip=0pt plus.2pt minus0.2pt

\subsection{Digitalization}
The \textbf{digitalization} stage involves transforming physical raw TCP into digital signals or codes, primarily as digital images~\cite{warwick2012digital}. 
These images could form a large corpus of TCP, comprising hundreds and thousands of artworks that can hardly be accessed physically in one place. 
Therefore, this stage presents technical requirements for \textbf{storage}, \textbf{retrieval}, and \textbf{restoration}. 

\textbf{Storage} requires storing a large amount of TCP digital images in the database. 
Many image-related techniques are developed to store and show paintings smoothly with different levels of detail~\cite{iiif}. 
However, there is little research targeting the storage of TCP digital images, which incorporates specific query and analytical requirements. 
For these large databases, information \textbf{retrieval} searches interested series of paintings efficiently from different dimensions. 
In addition to the metadata of TCP, such as authors and themes, content-based image retrieval utilizes similarities of paintings in terms of visual features~\cite{dong2020feature,hung2018study,wang2007image}. 
\textbf{Restoration} of TCP deals with the pigment fading and paper aging~\cite{guo2013image} in the digitalization stage.
Chen\etal\cite{chen2012simulating} simulated the aging and reverse-aging phenomena. 
Several studies apply image recovery techniques to restore the electronic forms of TCP in terms of stroke and brush~\cite{guo2013image} and colors~\cite{guo2013image,chen2012simulating,pei2006background,pei2004virtual,ding2012research}. 

\subsection{Interpretation} 
After obtaining the digital formats of original TCP, the next stage is to \textbf{interpret} them by applying computational approaches. 
The aspects of interpretation vary from micro-level analysis (\eg colors and objects~\cite{feng2022ipoet,chen2021poemgeneration}), meso-level analysis (\eg emotion extraction~\cite{feng2022ipoet}), to macro-level analysis (\eg the layout of the painting and white space analysis~\cite{fan2019evaluation}). 

In terms of techniques, the majority of the current work falls into two categories: local and global feature extraction. 
The former emphasizes the identification of relevant features, including handcrafted features (\eg brushwork~\cite{zhan2019,lai2016data} and color~\cite{guo2015novel,liu2014Classification}) and learned features with deep learning technologies (\eg object detection~\cite{meng2019elements,gu2019deep}). 
The latter focuses on the effectiveness of classification such as accuracy, in which most work utilizes learned features~\cite{liong2020automatic,meng2019elements}.
More technical details will be introduced in \autoref{sec:tech_tasks} and \autoref{sec:tech_feature}. 

\subsection{Creation} 
After obtaining useful features and insights from the analysis of existing paintings, another large proportion of studies focuses on the \textbf{creation} of new artworks. 
Image generation and video generation are two mainstream creation outcomes. 

\textbf{Image Generation.} 
The majority of studies focuses on generating new paintings, considering the unique characteristics in style and the artistic elements of TCP based on \textit{The Six Principles}. 
Style transfer takes existing inputs (\eg photos and prepared sketches) and output generated TCP artworks~\cite{li2021immersive,9413063,zhang2020detail}. 
Several studies also modify the content of the original input such as face replacement~\cite{li2021immersive}. 
In addition to generating from existing materials, a few works have experimented with creating artworks from scratch, including both content and style generation~\cite{xue2021end}. 

\textbf{Video Generation.} 
With the development of computer animation, a line of work has explored creating videos in a TCP style, aiming to express the \textit{Spirit Resonance} in \textit{The Six Principles} from a new perspective. 
They could be classified into two categories according to the input and animation entities. 
The first category is to input an existing video and apply video style transfer to transform the whole frame~\cite{lianginstance,zhang2009video}. 
The second category is to input a TCP and animate entities such as characters and animals on the painting~\cite{lianginstance,lai2016data,xu2006animating}.
In addition, 2.5D artworks~\cite{8419282,amati2010modeling} and video scribing showing the construction of TCP for educational purposes~\cite{8113507} are explored.

\subsection{Exhibition} 
As an essential type of artwork, the exhibition is a typical stage for promoting TCP to the general audience. 
In addition to the traditional approach of arranging items one by one in the museum, studies have been exploring interactive approaches to engage audiences in the exhibition. 
Existing work could be classified into three categories according to the interactive platforms, namely, touch-screen-based, XR-based, and others. 

\textbf{Touch screens} are commonly applied in today's museums. 
Hsieh\etal~\cite{hsieh2013viewing} presented an interactive tabletop for audiences to view detailed regions of TCP. 
Subramonyam\etal~\cite{subramonyam2015sigchi} developed an iPad application, ``Rice Paper," for artists to highlight and annotate key information of the TCP for the general public. 
They also printed a tangible booklet based on this application to guide audiences. 
CalliPaint~\cite{li2014writing} is a system that allows audiences or artists to conveniently create TCP-based digital artworks.

With the development of immersive devices, \textbf{XR (Mixed Reality)} has become a new creation platform for curators and artists. 
Several studies (\cite{jin2020reconstructing,zhao2020shadowplay2,yuan2016tunable}) reconstruct TCP in the VR (Virtual Reality) environment with 3D or 2.5D characters and objects. 
They are intended to provide an immersive experience that the static TCPs cannot fulfill.
Jin\etal~\cite{jin2022immersive}) evaluated the engagement of audiences when showing TCPs on the touch screen and the VR platform. 

In addition to the touch screen and XR, other \textbf{interactive installations} are also studied, such as using sensors to capture audiences' walking in the 3D space to generate Chinese Shanshui Paintings (\cite{le2019walking}) and applying real-time projector-camera system for audiences to interact with TCP (\cite{jin2007real}). 
\section{Computational Techniques}
\label{sec:tech}

In this section, we will discuss computational techniques that have been applied to TCP. Although TCP and natural images are similar in their modality as pictures, they differ in terms of technique used for interpretation and creation. We organize and elaborate the techniques from three perspectives as follows: 

\textbf{tasks} for which computational techniques are used (\autoref{sec:tech_tasks}), 
\textbf{extracted features} which the models use for these tasks (\autoref{sec:tech_feature}), 
and \textbf{rendering techniques} in which the model generate new paintings (\autoref{sec:tech_renderin}).
These categories of each aspect are listed in \hyperref[fig:techClasses]{Fig.~5}.

\begin{center}
\includegraphics[width=8.2cm]{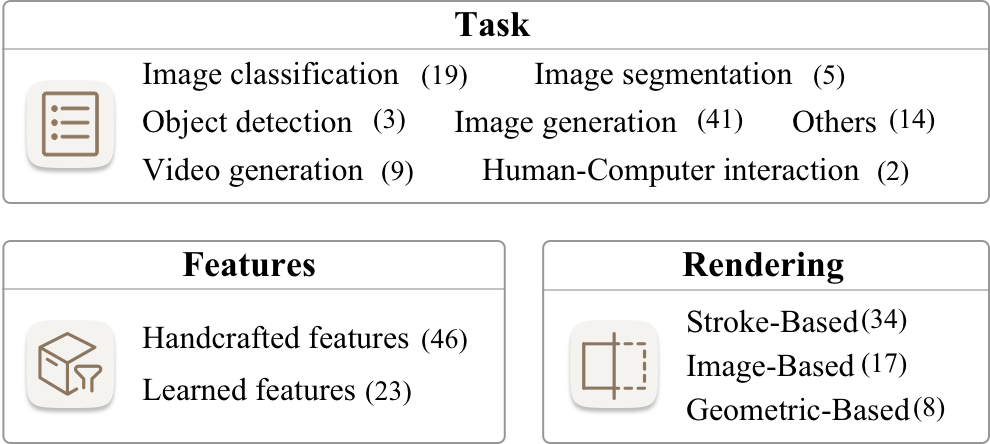}\\
\vspace{3mm}
\parbox[c]{8.3cm}{\footnotesize{Fig.5.~} Summary of computational techniques applied to TCP.}
\label{fig:techClasses}
\end{center}

\subsection{Tasks}\label{sec:tech_tasks}

Previous works on TCP mainly focus on tasks that resemble those in computer vision. Nevertheless, considering that TCP have distinct characteristics (as presented in \autoref{sec:background}) compared to natural images and videos, handling these tasks requires more TCP-specific designs and contributions.

\textbf{Image Classification.} TCP can be classified according to multiple attributes (\eg artists, painting techniques, and painting subjects). Annotating TCP with attributes can improve the retrieval experience and help understand the painting. Distinguishing TCP typically requires expert knowledge, which is time-consuming and expensive. Therefore, it is necessary to train automatic models for accurate TCP classification.

Many works~\cite{li2004studying,sheng2014recognition,liu2014Classification,sun2015brushstroke,sun2016monte,jiang2021mtffnet} classify TCP according to the artists in that the painting styles of different artists tend to be distinct. 
Specifically, for describing artists' painting styles, \cite{li2004studying} adopt a mixture of multiresolution hidden Markov models, and \cite{liu2014Classification} adopt various algorithms, such as Bayes, FLD, and SVM classifiers. \cite{sun2016monte} propose artistic descriptors with Monte Carlo Convex Hull for feature selection and use SVM for classification. 
Previous methods typically utilize traditional image processing techniques for classification. In contrast, \cite{sheng2014recognition}, \cite{sun2015brushstroke}, and \cite{jiang2021mtffnet} utilize MLPs or CNNs for distinguishing artists' styles.

TCP have two mainstream painting techniques, {\it Gongbi}~(a meticulous style, focusing on details) and {\it Xieyi}~(an ideographic style, expressing artists' feelings). \cite{jiang2019dct} apply discrete cosine transformation and CNNs for classifying {\it Gongbi} and {\it Xieyi} paintings, achieving promising performance.
Some other works focus on classifying painting subjects, including mountains-and-waters~(landscape), flowers-and-birds, and human figures. \cite{meng2018classification} apply a modified VGG~\cite{simonyanVery2015} network for painting subject classification, achieving 93.8\% accuracy. Since TCP is closely related to calligraphy, \cite{liang2010simple} distinguish TCP from calligraphy according to the Chinese characters' structures and the differences in image composition. \cite{li2020multi} propose an LSTM-based model to classify TCP into five categories: ancient trees, people, flowers-and-birds, Jiangnan water-bound town, and ink paintings. However, these categories overlap with each other in the TCP concept, which inevitably limits the model's generalization ability.

\textbf{Image Segmentation and Object Detection.} 
Image segmentation was studied in TCP with traditional morphological methods, yet recent neural network-based methods have not been explored. There are two reasons: (1) It is hard to collect large-scale training datasets of TCP, which require domain knowledge for annotation; (2) the object boundaries of TCP (specifically a key category, {\it Xieyi} painting) are hard to determine, as shown in \hyperref[fig:gongbixieyi]{Fig.~3}A. In spite of these difficulties, \cite{hu2015object} try to extract the foreground objects from a human-designed saliency map, which has a smaller dependency on the scale of data. Some works \cite{8419282,zhang2011multispectral,chen2012simulating} decompose the painting into multiple layers to obtain foreground objects or stroke segmentation. On the other hand, prefaces and postscripts are vital components of TCP~(as shown in \autoref{fig:tcp_w}A1), thus \cite{bao2010novel} propose a rule-based method to extract these scripts. 

Directly adopting natural image-tailored deep learning models for detecting objects (\eg figure, plant, flower) in TCP tends to have poor performance.~\cite{gu2019deep}. \cite{meng2019elements} utilize modified YOLOv3~\cite{redmon2018YOLOv3} and RetinaNet~\cite{lin2017Focal}, and \cite{gu2019deep} propose a modified RPN~\cite{ren2017Faster} by assembling low-level visual information and high-level semantic information. Apart from the categories that also appeared in natural images, a traditional Chinese painting may contain many seals that identify the owners and collectors in a long history, automatically detecting seals can greatly help understand the artwork~\cite{bao2009effective}. 

\textbf{Image Generation.} 
TCP have their own styles (\eg ink wash painting, white space) compared with other painting types, such as oil painting. Early works try to transfer a natural image into ink wash paintings by adjusting colors and textures~\cite{guo2015novel,dong2014real,zhang2011multispectral,yu2003image} based on tuned hyper-parameters. These early methods are learning-free, thus typically requiring tuning hyper-parameters for each image. Recent works~\cite{xue2021end,zhang2020detail,he2018chipgan,li2021immersive,9413063} have taken efforts to create TCP with Generative Adversarial Networks~\cite{goodfellow2014Generative} that transfer noises or natural images into paintings by adversarial training. Some other works~\cite{wu2018research} perform style transfer methods with CNNs by separately learning semantic information and styles from two source images and generating a blending image. These methods are machine learning models, requiring a number of training samples for learning millions of parameters. 

Apart from regarding the painting as a whole to generate, another group of works considers that Chinese paintings employ brush strokes and ink to depict objects on the paper or silk. Specifically, some works~\cite{xie2013artist,xu2005virtual,mi2004droplet,way2001synthesis,lee1999simulating,10.1145/15886.15911} model either the brush or various stroke shapes, pursuing better texture simulation of real brush strokes. With the specifically modeled brushes, users can draw Chinese paintings stroke by stroke on the screen, instead of drawing on papers with a real brush~\cite{yang2019easy,le2019walking,li2014writing}. Considering the characteristic of rice paper and silk, a large number of early works~\cite{liang2013image,xu2007generic,wang2007image,10.1145/1073204.1073221,guo2003nijimi,way2003physical,lee2001diffusion} model the ink diffusion on the rice paper and silk, seeking to improve the realism of paintings. 
In addition, previous methods focus on creating digital Chinese paintings. Yao\etal~\cite{yao2005painting} build a painting robot to handle the brushes and draw real paintings by simulating human actions. 

\textbf{Video Generation.} We divide the works on TCP video generation into three classes according to their targets: (1) displaying the painting process, (2) animating objects, and (3) natural video style transfer. For the first target, a few works~\cite{yang2013animating,8113507,yang2013animating} focus on the creation of Chinese painting, proposing to display the painting process of brush strokes for TCP. In this way, brush trajectory can be animated for both education and appreciation purposes. For the second target, some other works~\cite{liu2020animating,lai2016data,zhang2009video,xu2006animating} present methods to animate figures, flowers, and water for a vivid representation of elements in Chinese paintings. Zhao\etal~\cite{zhao2020shadowplay2} build a visualization system to build 2.5-dimensional stories about Chinese poetry, displayed by 360-degree videos, which is expected to provide an immersive appreciation of poetry in Chinese painting styles. For the third target, Liang\etal~\cite{lianginstance} display a deep learning-based multi-frame fusion framework to stylize natural videos with ink wash styles. In the process of transferring, object coherence between adjacent frames is specifically considered for semantic consistency. 

\textbf{Human-Computer Interaction.} Compared with videos, devices supporting human interactions typically have a well immersive experience to appreciate TCP. For instance, the 360-degree space in VR can better satisfy the demands of displaying handscroll. As building the scenery on the virtual reality platform is labor-intensive and expensive, existing works only focus on a single painting. Specifically, Yuan\etal~\cite{yuan2016tunable} reconstruct a painting ``Listening to a Guqin'' in the mode of virtual reality, and Jin\etal~\cite{jin2020reconstructing} build the 3D scene of the painting ``Spring Morning in the Han Palace'' using a head-mounted platform. Moreover, the immersive multi-touch tabletop is also a promising interactive method for facilitating learning and appreciating TCP~\cite{jin2022immersive,subramonyam2015sigchi,hsieh2013viewing}. Ma\etal~\cite{ma2012annotating} embed the audio explanation into the local area of the Chinese painting, thus enabling the user to move the focus to get the audio explanation of the corresponding area while enjoying the painting. Jin\etal~\cite{jin2007real} develop a real-time projector-camera system that allows users to interact with Chinese ink cartoons (\eg interacting with water can create ripples). 

\textbf{Others.} Apart from the discussed tasks in CV and HCI above, there are various tasks involving TCP, such as color recovery~\cite{ding2012research}, poet generation from TCP~\cite{feng2022ipoet,chen2021poemgeneration}, Chinese painting retrieval~\cite{dong2020feature,hung2018study,zhang2004modelling}, white space understanding~\cite{fan2019evaluation}, and digital image enhancement~\cite{guo2013image,chen2012simulating,pei2006background,pei2004virtual}.

\subsection{Feature Extraction}\label{sec:tech_feature}
There are abundant features in TCP which distinguish them from many other painting genres.
To help users analyse and learn from TCP, many researches have extracted features for downstream tasks, such as painting classification and creation.
We summarized the features of TCP into two categories, \ie handcrafted features and learned features, according to the methodology of feature extraction.

\textbf{Handcrafted features.}
The handcrafted features are extracted by rule-based methods and reflect the specific aspects of TCP.

\emph{Brushwork} is an important feature in depicting the bone method of the paintings. 
Typically, the TCP are created with brushes dipped in ink, and the ink permeates through the rice paper, creating the unique shape of the brush strokes.
To automatically generate the TCP, a wide range of studies \cite{zhan2019,lai2016data,dong2014real,xu2012stroke,amati2010modeling,bai2009chinese,xu2007generic,yao2005painting} focusing on simulating the diffusion effect of color ink.
Wang\etal~\cite{wang2007image} proposed a physically-based model with texture synthesis method to simulate the color ink diffusion.
Chu\etal~\cite{10.1145/1073204.1073221} introduced a fluid flow model to calculate the percolation in the paper.
In addition, the brushwork is related to the visual complexity of the paintings. Dense thin strokes can increase the complexity while sparse thick strokes lower the complexity.
Fan\etal~\cite{fan2017visual} measured stroke thickness based on the calculation of color change.
Combining the analysis of stroke structures with the ink dispersion densities and placement densities, \cite{lai2016data} generated animations for water flow in the TCP according to the stroke pattern groups of the flow field.

\emph{Color} is another significant factor and implies the types of the TCP style \cite{guo2015novel,liu2014Classification,chen2012simulating,feng2022ipoet,lu2008content,wang2007image}.
Liu\etal~\cite{liu2014Classification} extracted the color information of the paintings by calculating the mean and variance values of the image pixels, and used them to support painting classification tasks.
Color can also be used in painting retrieval \cite{hung2018study}, painting style modeling \cite{feng2022ipoet}, and painting enhancement \cite{chen2012simulating}.
Over time, ancient Chinese paintings have faded and aged, requiring human restoration. 
Pei\etal~\cite{pei2004virtual,pei2006background} design color enhancement schemes to improve the image contrast, making the paintings more vivid and bright.

\emph{Objects}, such as the scenery in the paintings, are the basic elements of the painting composition and contain semantic information. Ding\etal~\cite{zhang2004modelling} extract objects by labeling pixels according to their connectivity in a pre-processed image, and use them for image retrieval. 
Feng\etal~\cite{feng2022ipoet} extract the objects in the TCP and use them to describe the painting content and create the painting poetry.
Zhao\etal~\cite{zhao2020shadowplay2} built a TCP style image repository for basic objects, supporting users to create immersive videos for poetry appreciation.

\emph{Scripts} are written in the empty space of the paintings and serve as complementary expression of creators' artistic ideas.
Bao\etal~\cite{bao2010novel} automatically identify and extract the scripts from the paintings according to their colors and regions.
Several studies also focus on other feature of the TCP, such as the white space~\cite{fan2017visual}, composition~\cite{sun2016monte}, and seal images~\cite{bao2009effective}.

\textbf{Learned features.}
With the fast development of deep learning technology, many studies have introduced deep learning models to learn the features of TCP.
Based on the labeled data of TCP, supervised learning methods (\eg CNN~\cite{krizhevsky2017imagenet}, VGG-16~\cite{simonyan2014very}, and YOLOv3~\cite{redmon2018YOLOv3}) are applied in object detection \cite{meng2019elements,gu2019deep,feng2022ipoet}, image classification \cite{liong2020automatic,meng2019elements,meng2018classification}.
As the stylistic features of the TCP are unique from other paintings, it is valuable to learn the stylistic features to transfer neural images into TCP.
A range of studies \cite{xue2021end,lianginstance,9413063,he2018chipgan} focus on capturing the stylistic features of TCP with adversarial training, a classical learning strategy in unsupervised learning.
In contrast, Li\etal~\cite{li2020multi} introduce weakly-supervised learning for semantic classification in the scenario with limited number of training images.
\subsection{Rendering} \label{sec:tech_renderin}
TCP rendering techniques are adopted in the process of image and animation generation. According to the focus of used techniques in rendering, we classify the TCP rendering methods into three classes: stroke-based, image-based, and geometry-based. 

\noindent\paragraph{Stroke-Based Rendering.} Users can create TCP through simulated paint brushes to be rendered on a digital canvas. Strassmann\etal~\cite{10.1145/15886.15911} propose a realistic model of painting including Brush (a series of bristles with ink supply and positions), Stroke (a set of parameters like position and pressure), Dip (a procedure to assign states to each bristle of the brush), and Paper (the carrier of ink as it comes off the brush). With the four elements, one can build an interactive or automatic painting software on the computer. 

Typically, there are two types of methods to generate brush strokes. The first method models the stroke boundaries with B'{e}zier or B-spline curves and then fills the closed curves with designed textures~\cite{chua1990bezier,nishita1993display}. The other method directly models the two-dimensional brush, such as the work of Strassmann\etal~\cite{10.1145/15886.15911}. However, the brush bristles are visually fixed in shape, users cannot apply such an e-brush with their realistic painting skills. Some works~\cite{lee1999simulating,yeh2002effects} develop ``soft'' brushes in which the shape of bristle bundle varies in response to the forces given by users. Furthermore, Xu\etal~\cite{xu2005virtual} model brushes with writing primitives (a bundle of hair bristles), instead of each single brush bristle, to improve the simulating realism. To further simplify the model complexity, Bai\etal~\cite{bai2009chinese} propose a geometry model to simulate the entire brushes, instead of large amount of bristles. A dynamic model is also introduced to simulate the brush deformation under the internal and external forces. 
Previous methods focus on modeling general brush strokes, some methods propose tailored algorithms to model specific object shapes and textures such as rocks~\cite{way2001synthesis}, trees~\cite{way2002synthesis}, bamboos~\cite{yao2005painting}, and water~\cite{zhang2009video}. Instead of small brush strokes, Fu\etal~\cite{8419282} decompose the painting into image layers with each layer representing a class of specific strokes. With these stroke layers, they can create a new high relief, an art form between 3D sculpture and 2D painting. 

For animation generation, Xu\etal~\cite{xu2006animating} build a brush stroke library obtained from painting experts, and animate the paintings by decomposing them into brush strokes and changing these strokes. Zhang\etal~\cite{zhang2009video} create running water animations with a novel proposed painting structure generation method, which is used to estimate water flow line positions. Previous methods create brush trajectory relying on manual inputs, Yang\etal~\cite{yang2013animating} automatically estimate such trajectory from the paintings by modeling the brush footprint. Considering the automation process of extracting brush trajectory, some methods~\cite{8113507} try to reconstruct the drawing process by estimating and animating the drawing order of brush strokes.

Apart from the brush strokes, ink diffusion in paper fibers structure is also a critical characteristic, which has been studies in literature~\cite{kunii1995diffusion,zhang1999simple,lee2001diffusion,yu2002model,guo2003nijimi,10.1145/1073204.1073221,wang2007image,xu2007generic,liang2013image}. Specifically, Kunii\etal~\cite{kunii1995diffusion} propose a multidimensional diffusion model to simulate the ink density distribution as in real paper. Some works~\cite{lee2001diffusion,way2003physical,way2006computer,wang2007image} further simulate the ink of brush strokes on various types of paper based on physical-based models. Considering the potential blending of multiple strokes, Yeh\etal~\cite{yeh2002effects} and Yu\etal~\cite{yu2002model} build the ink diffusion simulation model with multi-layered structures of brush and paper. Chu\etal~\cite{chu2004real} develop a system for creating painting with more complicated ink diffusion effects based on lattice Boltzmann equation, and accelerate the algorithm for real-time process by utilizing both CPU and GPU. 

\noindent\paragraph{Image-Based Rendering.} Previous methods mainly focus on modeling the brush strokes and ink diffusion for interactive painting creation. From another technical route, one can directly synthesize Chinese painting from existing images. For instance, Yu\etal~\cite{yu2003image} propose a framework for image-based painting synthesis. Specifically, the authors build a brush stroke texture primitive collection, and map those texture primitives to a constructed mask image (named as control picture in \cite{yu2003image}). Apart from blending strokes, some works propose style transfer methods to transform natural pictures to paintings, involving handcrafted feature-based flow~\cite{liang2013image,dong2014real,guo2015novel} or deep neural networks. Typically, deep neural network based style transfer methods~\cite{wu2018research,he2018chipgan,zhang2020detail,9413063,li2021immersive,xue2021end} are data-driven, characterised by training the model with tailored training program and large scale data sets. For instance, ChipGAN~\cite{he2018chipgan} consists of a generator and a discriminator, where the generator is trained to transfer photos into paintings while the discriminator is trained to discriminate the generated paintings and real paintings. Meanwhile, ChipGAN requires thousands of images for training the model due to the large scale trainable parameters. 

For animating Chinese paintings, Liu\etal~\cite{liu2020animating} propose a sample point processing method to preserve the style of brush strokes and determine control bones, and a skeleton-based deformation method for animation generation. Liang\etal~\cite{lianginstance} leverage deep neural networks for transferring natural videos into ink wash painting-style videos. In order to enhance temporal consistency between video frames, the authors introduce multi-frame fusion and implement instance-aware style transfer, which help generate paintings with proper white space. 

\noindent\paragraph{Geometry-Based Rendering.} Stroke-based and image-based rendering are both from the view of computer vision. Instead, geometry-based methods adopt the view of computer graphics. Chan\etal~\cite{chan2002two} decompose the brush stroke-like features into layers of procedural shaders, and then mix different layers to construct desired effects in 3D models. Some works~\cite{way2002synthesis,yeh2002non,xu2012stroke} synthesis objects with ink painting styles by building polygonal models or extracting silhouette, and then mapping specific textures on the models. Amati\etal~\cite{amati2010modeling} develop a webcam-based system to capture the process of user drawing plants and build corresponding 2.5-dimensional digital models. Different from previous methods, Shi~\cite{shi2017generative} build landscape paintings from a 3-dimensional city model by generating mountains from buildings and assigning ink painting styles.

\section{Challenges and Oppurtunities}
\label{sec:challenge}

\subsection{Lack of large-scale and high-quality datasets.}
The lack of large and high-quality open-access datasets is a crucial reason hindering the further development of Traditional Chinese Painting research.
According to our paper, some works have announced that they have produced a few Chinese painting datasets~\cite{hung2018study, xue2021end, 9413063, liong2020automatic, dong2020feature}. For example, Liong et al.\cite{liong2020automatic} constructed an unlabeled dataset containing more than 1,000 Chinese paintings, and Dong et al.\cite{dong2020feature} collected a labeled dataset. However, these datasets are limited in size and have not been made open-source.
Building a large and high-quality Chinese painting dataset faces several challenges, including:

\begin{itemize}
\setlength{\itemsep}{0pt}
    \item \textit{Data availability}. As most Chinese paintings are held in museums and private collections all over the world, there is a problem of copyright ownership. It is particularly difficult to collect online resources of Chinese paintings.
    \item \textit{Data quality}. Many famous Chinese paintings are large in size, rich in details, and difficult to preserve, leading to the high cost of digitizing Chinese painting and a high technical barrier for generating high-definition pictures.
    \item \textit{Data diversity}. Chinese paintings contain relatively independent items, such as colophons and seals that can be used for analyzing historical events, collection paths, etc. However, only a few articles discussed the extraction of colophons and seals~\cite{bao2010novel,bao2009effective} without further exploration.
    \item \textit{Data annotation}. Due to the domain professionalism, annotating Chinese painting data requires high-level expertise, especially for systematic annotations based on the \textit{Six Principles of Painting}, which can be extremely expensive.
\end{itemize}

\subsection{Insufficient consideration on the TCP uniqueness.}
The analysis of TCP research tendencies based on the \textit{Six Principles of Painting} in \autoref{sec:background} reveals that most articles focus on \textit{Bone Manner, Structural Use of the Brush} (36/92), and \textit{Conform with the Objects, Obtain their Likeness} (24/92). This is mainly because the recognition, segmentation, and classification of strokes and objects in paintings can be formulated into CV tasks and solved with mature CV models.
However, researchers have paid little attention to the unique stroke system in traditional Chinese painting (3/92)~\cite{way2002synthesis,way2001synthesis,zhang1999simple}. For example, there are eighteen unique drawing methods in TCP techniques for depicting portraits (\autoref{fig:tcp_w}), and different wrinkling techniques for depicting mountains and rocks (\autoref{tbl:tcp_oil}). A detailed analysis of the stroke system sheds lights on the painter's style and the mentoring relationships between painters. Therefore, it is an issue that deserves attention in future computer fields.

Moreover, the \textit{Place and Position} aspect is not given as much attention in the current work (4/92). According to experts, the composition of Chinese paintings is crucial. Painters often use white space to convey mood, inscriptions, and seals to balance the picture's composition. Therefore, a computer-based systematic examination of the composition of Chinese paintings can aid specialists in understanding the compositional traits of paintings across time.

Regarding the study of \textit{Movement of Life} (9/92), it is important to note that in recent years, the fusion of AR and VR technology into TCP has risen, allowing audiences to experience Chinese painting from a new perspective. The development of new technology has increased the opportunities for the study and presentation of TCP, but more work of this type is required, and it may be reinforced in the future.

\subsection{Disregard for the data-linking in TCP analysis.}
Cultural heritage has various types of data, including paintings, ancient books, sculptures, architecture, and more. 
As a form of cultural heritage, TCP has garnered attention in recent years. 
However, most current research has focused solely on TCP data, and rarely combines other datasets for cross-analysis. 
From a historical research perspective, TCP collections represent only a snapshot of a certain time period. It is possible that snapshots of different artifacts describe the same social landscape. To gain a more comprehensive historical understanding, different collections of cultural relics should be viewed together. 
Therefore, it is worthwhile to pay attention to how to integrate different data related to paintings in order to restore a more accurate historical picture.

On the other hand, the use of multi-modal data to construct deep learning models is a growing trend. Multi-modal data enables better feature representation construction in the latent space, which improves the fusion of textual, visual, and other forms of information like videos and knowledge graphs. This ultimately strengthens the model's performance and enhances its generalization capacity.

\subsection{Insufficient exploration of ML methods and large models on TCP.}
TCP image data has unique characteristics compared with natural images, such as cross-domain, few annotated training samples, imbalanced classes, and variable sizes of objects. Advanced ML methods have taken profound discussions on related topics such as transfer learning, domain adaptation, domain generalization, few-shot learning, and learning with long-tailed data distribution. Therefore, applying these advanced methods to TCP can promote Chinese painting analysis from a computational view. Meanwhile, these methods can effectively reduce the demand for data annotations and alleviate the burden of collecting large-scale and high-quality annotated datasets.

In addition, large language models (\eg GPT-4~\cite{openai2023gpt4}) and large vision models (\eg CLIP~\cite{radford2021Learninga}, Stable-Diffusion~\cite{rombach2022HighResolution}) are becoming the new foundations of advanced research. Current models are not specifically adapted to TCP data, tending to generate images that ignore the Six Principles of Paintings, as well as textual descriptions that often lack detail and do not capture the essence of the painting. It is inevitable that unimodal or multimodal large models will be adopted in traditional Chinese painting research.

\subsection{Inadequate applications for artwork creation and promotion.} 
Although a line of work has explored the generation of TCP-styled paintings and videos, the quality of these AI-generated artworks could be doubtful. 
Involving artists in the creation process with a semi-automated creation style would be a promising direction in the future. 
In addition, advanced display and interaction techniques (\eg immersive techniques) should be applied to promote TCP to the general public. New storytelling approaches should also be constructed to enhance the appreciation and understanding of TCP.  
\section{Conclusion}
In this paper, we analysed 92 papers to understand the current applications of computer technology in TCP. In collaboration with experts, we examined these papers from three perspectives (\ie subject, framework, and technique). On top of them, we addressed the potential research directions for utilizing computer technologies in TCP. This paper can aid with future study by serving as a useful introduction to the area of computer methods used with TCP.

\end{multicols}
\label{last-page}
\end{document}